\documentclass[a4paper,fleqn]{cas-sc}

\usepackage[square,numbers]{natbib}
\usepackage{amsmath}
\usepackage{graphicx}
\usepackage{subcaption}
\usepackage{float}
\usepackage{placeins}
\usepackage{lipsum}
\usepackage{soul,color}
\usepackage{subcaption}

\numberwithin{equation}{section}

\usepackage[pagewise]{lineno}

\def\tsc#1{\csdef{#1}{\textsc{\lowercase{#1}}\xspace}}
\tsc{WGM}
\tsc{QE}
\tsc{EP}
\tsc{PMS}
\tsc{BEC}
\tsc{DE}

\begin{document}

\let\WriteBookmarks\relax
\def\floatpagepagefraction{1}
\def\textpagefraction{.001}
\shorttitle{A Framework for Pedestrian Sub-classification and Arrival Time Prediction at Signalized Intersection Using Preprocessed Lidar Data}
\shortauthors{T. Lin, Z. Jin, S. Choi and H. Yeo}

\title [mode = title]{A Framework for Pedestrian Sub-classification and Arrival Time Prediction at Signalized Intersection Using Preprocessed Lidar Data}




\author[1]{Tengfeng Lin}[orcid=0000-0003-1397-6603]
\ead{tengfenglin@kaist.ac.kr}
\credit{Conceptualization,
Methodology,
Software,
Validation,
Visualization,
Formal analysis,
Writing - Original Draft}

\author[1]{Zhixiong Jin}[orcid=0000-0002-1370-781X]
\ead{iziz56@kaist.ac.kr}
\credit{Conceptualization,
Methodology,
Software,
Visualization,
Formal analysis,
Writing - Original Draft}

\author[2]{Seongjin Choi}[orcid=0000-0001-7140-537X]
\ead{seongjin.choi@mail.mcgill.ca}
\credit{Conceptualization,
Methodology,
Formal Analysis,
Visualization,
Writing - Review \& Editing}


\author[1]{Hwasoo Yeo}[orcid=0000-0002-2684-0978]
\cormark[1]
\ead{hwasoo@kaist.ac.kr}
\credit{Conceptualization, Supervision, Project administration, Writing - Review \& Editing}

\address[1]{Department of Civil and Environmental Engineering, Korea Advanced Institute of Science and Technology, 291 Daehak-ro, Yuseong-gu, Daejeon, 34141, Republic of Korea}
\address[2]{Department of Civil Engineering, McGill University, 817 Sherbrooke Street West, Montreal, Quebec H3A 0C3, Canada}

\cortext[cor1]{Corresponding author}

\begin{abstract}
The mortality rate for pedestrians using wheelchairs was 36\% higher than the overall population pedestrian mortality rate. However, there is no data to clarify the pedestrians' categories in both fatal and nonfatal accidents, since police reports often do not keep a record of whether a victim was using a wheelchair or has a disability. Currently, real-time detection of vulnerable road users using advanced traffic sensors installed at the infrastructure side has a great potential to significantly improve traffic safety at the intersection. In this research, we develop a systematic framework with a combination of machine learning and deep learning models to distinguish disabled people from normal walk pedestrians and predict the time needed to reach the next side of the intersection. The proposed framework shows high performance both at vulnerable user classification and arrival time prediction accuracy. 
\end{abstract}


\maketitle

\section{Introduction}
Road traffic safety has ranked among the most pressing transportation concerns around the world; approximately 1.35 million people are killed and 50 million injured in traffic accidents each year \cite{peden2005global}. In particular, while the total number of traffic deaths in the US has been on the decline, the number of deaths by pedestrians has been on the rise, which accounted for about 53\% between 2009 and 2018 \cite{bernhardt2021analysis,national20192018}.  More than half of the road traffic deaths among pedestrians are related to vulnerable road users (VRUs): wheelchair, cyclists, and other \cite{timmurphy.org}. From 2006 to 2012, the mortality rate for pedestrians using wheelchairs was 36\% higher than the overall pedestrian mortality rate.
In addition, almost half of the fatal crashes occurred in intersections, and 38.7\% of intersection crashes occurred at locations without traffic control devices. Among intersection crashes, 47.5\% involved wheelchair users in a crosswalk; no crosswalk was available for 18.3\%. 
According to government statistics, each year in the U.S, about 5,000 pedestrians in and out of wheelchairs are killed and another 76,000 are injured in crashes on public roads \cite{Traffic_Safety}. As a result of the aforementioned facts, it is critical to protect vulnerable road users, particularly when they are crossing intersections.




While this is devastating, it is more devastating to know that there is no data on the pedestrians' categories in both fatal and nonfatal accidents, since police reports often do not keep a record of whether a victim was using a wheelchair or has a disability. Wheelchair Users or disabled people, currently rely on pre-programmed signal timing and pedestrian phases to safely cross a signal-controlled intersection. However, in real life, these pedestrians are hard to cross the intersection within original the green signal time due to low moving speed. To solve this issue, these VRUs should benefit from more adequate and proactive protection that recognizes, differentiates, distinguishes a walking pedestrian from a handicapped one, predicts and compares the time needed to reach the next side of the intersection with the current left green time, and implements real-time adjustment on the current phase to provide extra time for VRUs to cross the road. In other words, it is necessary to predict the left time or arrival time for VRUs to cross the streets in order to decide if the green signal time is enlarged or not. 

The main purpose of this paper is to develop an accurate and novel method for sub-classification, predict the time needed to cross the street, and justify whether the pedestrian can safely cross within the allocated pedestrian signal time. Therefore, in this paper, we propose a framework for pedestrian sub-classification and arrival-time prediction at the signalized intersection. The framework mainly consists of two models: pedestrian sub-classification and arrival-time prediction.     
To sub-classify the pedestrians at higher risk, we first develop a high-accurate machine learning-based classification model, which utilizes the static size feature of the test site and motion of pedestrian features. 
To predict the time needed to cross the street (or arrival-time prediction) and justify whether the pedestrian can safely cross within the allocated pedestrian signal time, we implement four different deep learning models with a data augmentation method for solving data sparsity problems. The proposed framework can benefit to decide the duration of green time, which can make a more safe intersection for pedestrians, especially disabled people.   


\begin{figure}[t]
  \centering
  \includegraphics[width=0.7\textwidth]{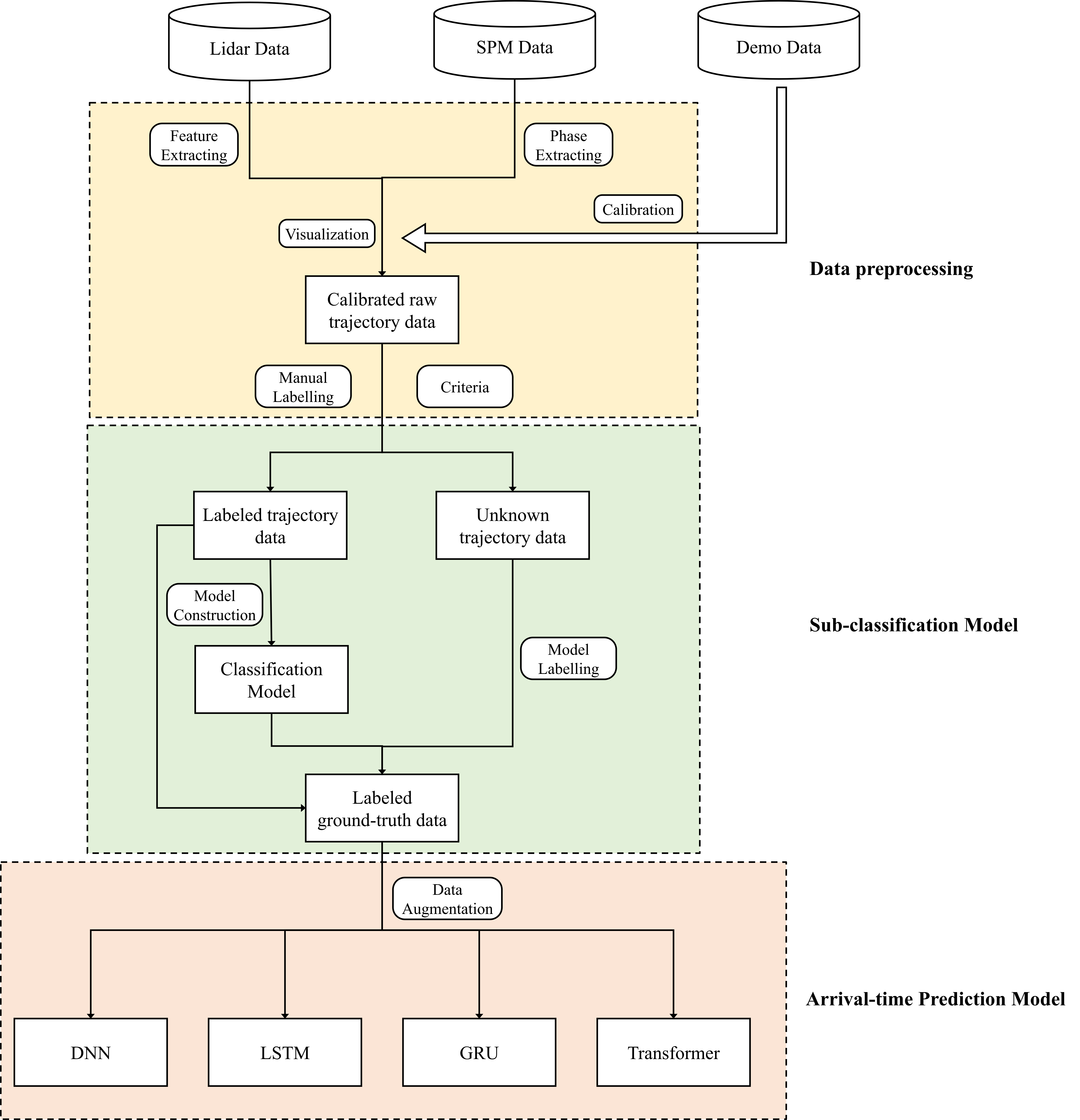}
  \caption{Framework for pedestrian sub-classification and arrival time prediction}\label{fig:methodology}
\end{figure}

\section{Methodology}
The proposed framework consists of two main models: $(1)$ Pedestrian sub-classification model and $(2)$ Pedestrian arrival-time prediction model. 
For the first model explanation, we focus on how to develop a high-accurate classification model with limited ground-truth data efficiently. We first sub-classify the pedestrians into three different categories: Normal Pedestrians, Wheelchair Users, and Unknown based on the specified four criteria. Then, Principal Component Analysis (PCA) is used to verify the reliability of the proposed criteria and to find the differences between each sub-classified dataset. Then, we use Support Vector Machine (SVM) to build a sub-classification model based on our sub-classified dataset.     
Then in the second model explanation, we use four deep learning models: Deep Neural Network (DNN), Long Short Term Memory (LSTM), Gated Recurrent Unit (GRU), and Transformer, which are suitable in time-series data processing, to find the best model for pedestrian arrival-time prediction. To overcome the data sparsity problems in model training, we also propose a rule-based data augmentation method in this paper. The proposed framework is shown in Figure~\ref{fig:methodology}. The whole framework consists of three parts: data preprocessing, sub-classification model development, and arrival-time prediction model development. In the next section, we will introduce each part in detail.  
%


\subsection{Data Prepossessing}
This paper focuses on how to sub-classify the pedestrian at higher risk and predict the time needed to cross the street. Therefore, it is necessary to extract the pedestrian-only trajectory accurately. However, due to the detection deficiencies and environment noise, there are unavoidable errors in our raw data. Therefore, we set three basic rule to do the data prepossessing.

\begin{itemize}
    \item \textbf{(Rule 1)}: For the trajectory with the same ID, if the total amount of label number 2 (pedestrian) comprises 50\% of the data, we consider it as pedestrian trajectory. 
    
    \item \textbf{(Rule 2)}: For the trajectory with the same Id, we only consider the trajectory which appears on the cross-section.

    \item \textbf{(Rule 3)}: For the trajectory with the same Id, we only consider the trajectory which contains more than 10 points. 

\end{itemize}

\subsubsection{Calibration and Background Matching}
\begin{figure}[t]
  \centering
  \includegraphics[width=0.7\textwidth]{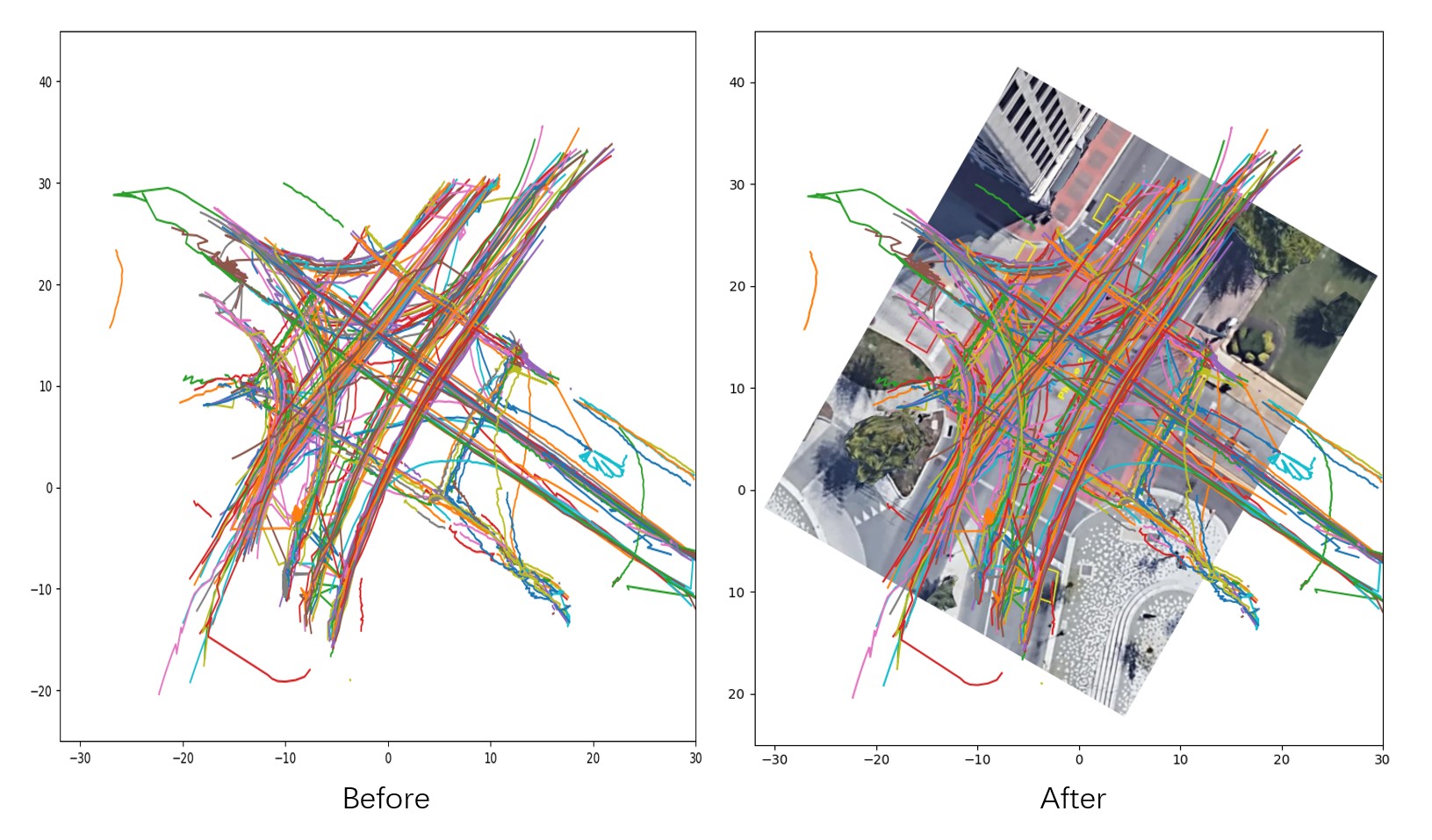}
  \caption{Background Matching}\label{fig:calibration}
\end{figure}

To better understand the location information of the trajectories, it is necessary to do the calibration. In the demo dataset, an intuitive demo video is provided. We manually calibrate the location of the pedestrian trajectories referred to in the demo video for further visualization. The results are shown in Figure~\ref{fig:calibration}.

\subsubsection{Visualization}
\begin{figure}[t]
  \centering
  \includegraphics[width=1\textwidth]{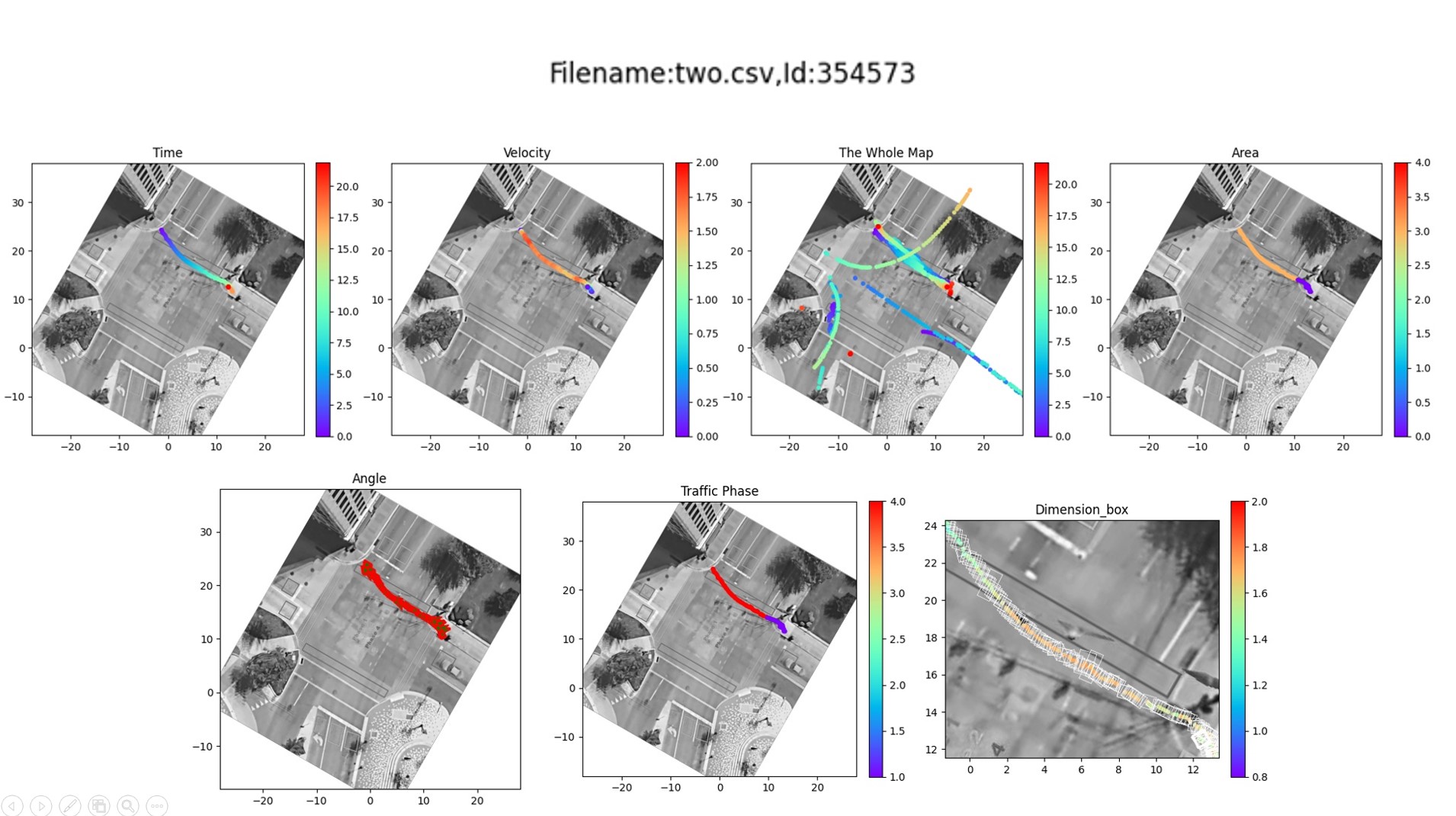}
  \caption{Visualization trajectory of Id 354573}\label{fig:visualiztion}
\end{figure}

To further determine sub-classification criteria, we visualize the pedestrian trajectories with different features in the input data. In Figure~\ref{fig:visualiztion}, the visualization step presents six different features for the certain trajectory. The stated features are Total Cost Time, Yaw angle, velocity, size information, surrounding environment situation, and the relationship between velocity and Total Cost Time. 

\subsubsection{Criteria of Sub-Classification and Manual Labeling }
\begin{figure}[!ht]
  \centering
  \includegraphics[width=1\textwidth]{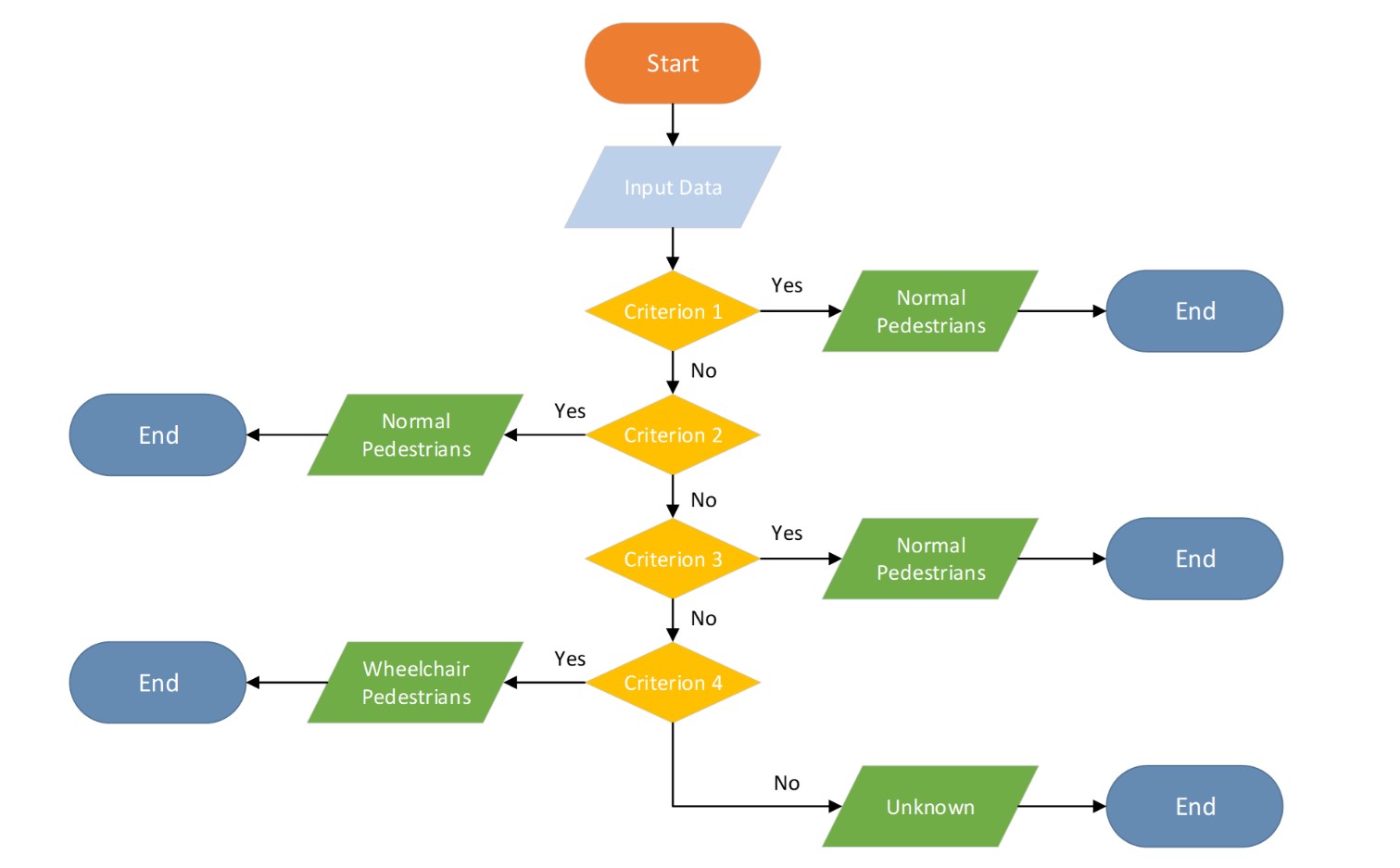}
  \caption{Criterion of Sub-classification}\label{fig:subclassification}
\end{figure}

In the raw data, the classification results are not provided. However, it is necessary to classify the trajectories with different labels to develop the sub-classification model. In this paper, we propose three criteria: Normal Pedestrians, Wheelchair Users, and Unknown. The related criteria of sub-classification are shown in Figure~ \ref{fig:subclassification}: 


\begin{itemize}
\item \textbf{(Criterion 1)}: For the pedestrians crossing the street, if they accelerate before phase change, we can consider them as Normal Pedestrians. 

\item \textbf{(Criterion 2)}:For the pedestrians crossing the street, if the mean velocity is higher than 1.5m/s, we consider them as Normal pedestrians. According to \cite{coutinho2013determination}, the average moving velocity of wheelchair propulsion is 70.4 $m/min^{-1}$ (±21.1).

\item \textbf{(Criterion 3)}:For the pedestrians with a high variance of yaw angle ,they are classified as normal pedestrians 

\item \textbf{(Criterion 4)}:For the pedestrians who does not meet criterion 1-3, also the heights are  between 1.1m to 1.5m and the shape of the pedestrian is like a square, we classify them into the Wheelchair Users \cite{richter2001effect}.  

If the trajectories do not meet the criterion 1-4, we also classify them into Unknown.

\end{itemize}

After six criterion filtering, we get 96 pedestrian trajectories with 53 Normal Pedestrians, 32 Wheelchair Users, and 11 Unknown trajectories. To further develop the classification model and classify the unknown data, in the next section, we will introduce the sub-classification model.

\subsection{Sub-classification Model}
In sub-classification model development, we first have to classify if there is a significant difference between each categorized dataset. After finding it, we can develop a suitable classification model. Therefore, in the paper, Principal Component Analysis (PCA) is implemented first to find the difference between each dataset, and then Support Vector Machine (SVM) is utilized to develop a high-performing classification model.

\subsubsection{Principal Component Analysis(PCA)}
Principal component analysis (PCA) is the process of computing the principal components and using them to perform a change of basis on the data, which is shown in Figure~\ref{fig:PCA_SVM mechanisms} (a). Sometimes the PCA is only the first few principal components and ignores the rest \cite{wold1987principal}. The most important use of PCA is to decrease the number of input features and compute the principal components among the features. The reason to reduce the variables is to observe trends, jumps, and clusters of the original dataset. In this paper, the main purpose of using PCA is to find if there are significant differences between each categorized dataset, which is classified by our proposed criteria in the data preprocessing step. After applying PCA, we can determine if it is possible to implement SVM in sub-classification model development. In other words, if we cannot find any differences in the dataset, we have to redefine our proposed criteria for pedestrian subclassification.

\subsubsection{Support Vector Machine(SVM)} 
\begin{figure}[!ht]
  \centering
  \includegraphics[width=0.8\textwidth]{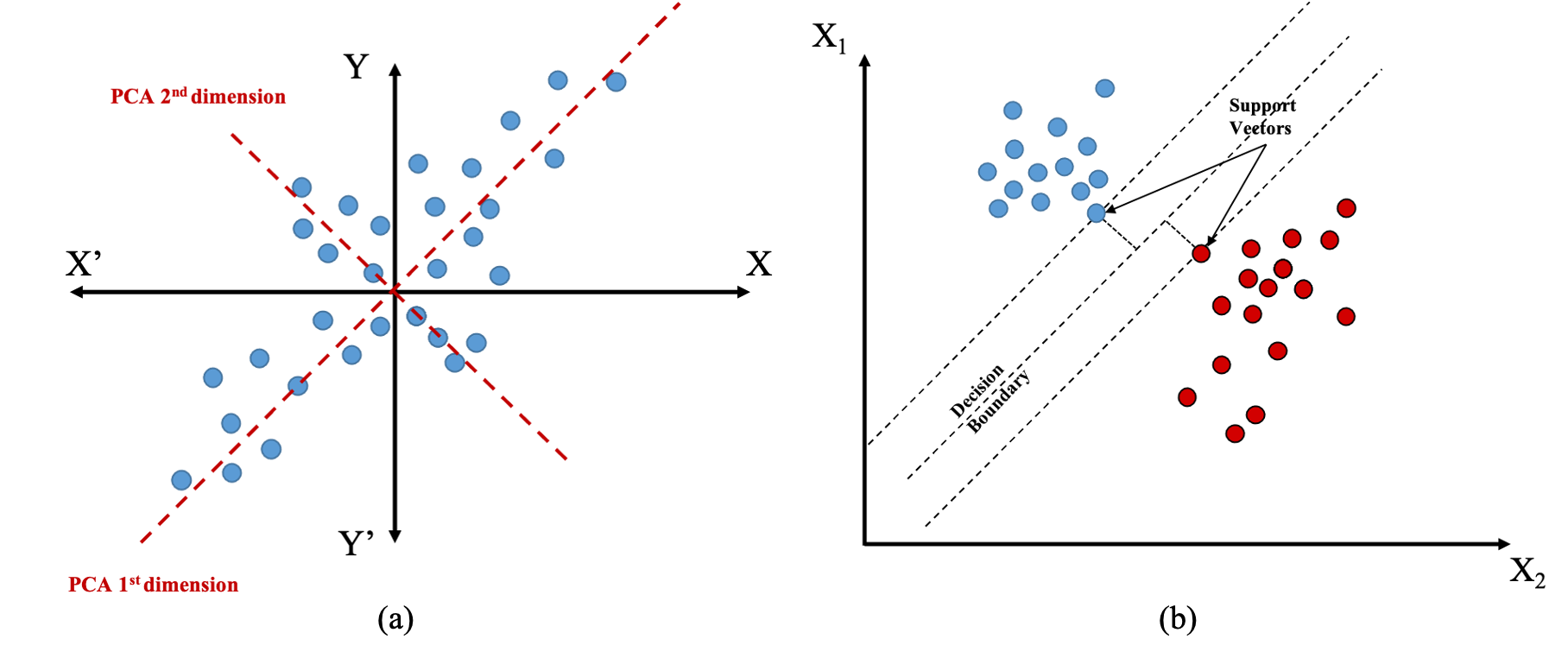}
  \caption{(a) PCA mechanisms (b) SVM mechanisms}\label{fig:PCA_SVM mechanisms}
\end{figure}

In machine learning, the Support Vector Machine (SVM) is a supervised learning model with associated learning algorithms that analyze data for classification and regression. \cite{noble2006support}. It is a linear model that can solve linear and non-linear problems and show good performance in many practical problems. The main mechanism of SVM is that it creates a line or a hyperplane which separates the data into several classes as shown in Figure~\ref{fig:PCA_SVM mechanisms} (b). In this paper, with the labeled data based on four proposed criteria, training the SVM for classification task aims to explore the implicit relationship between the feature and sub-classes,  which contains more implicit classification rules than the manual criterion methods. With the trained SVM model, the previous pedestrian with 'Unknown' labeled is redefined by the SVM model.

\subsection{Arrival-time Prediction Model}

In developing pedestrian arrival-time prediction model, there are two main challenges. One of the biggest challenges is building a high-performing model with limited ground-truth data. In this research, we use data augmentation method to solve the data sparsity problem. The other challenge is related with choosing a suitable deep learning model for model development. Therefore, to select a suitable model for arrival-time prediction, we test four common deep learning models: Deep Neural Network (DNN), Long Short Term Memory (LSTM), Gated Recurrent Unit (GRU) and Transformer in this research. In the following sections, we will explain how we solve the stated problems in detail. 

\subsubsection{Data Augmentation}
In deep learning model development, the augmentation methods are used to solve the data sparsity problem. The data augmentation methods in trajectory generation are mainly divided by rule-based and data-based methods \cite{jin2021transformer}. The rule-based methods are defined as generating trajectories based on pre-defined rules. Different from rule-based method for trajectory generation, the data-based methods generate trajectories based on the know data such as data duplication \cite{travis2008trajectory}, Markov chain \cite{chen2011discovering}, Generative Adversarial Network (GAN) \cite{wang2021large}, Generative Adversarial Imitation Learning \cite{choi2021trajgail}. However, the data-based methods need a sufficient number of ground-truth data for augmenting, which is not suitable for our task. Therefore, in this paper, we choose a rule-based model for data augmentation to solve the problem. The main idea of the proposed data augmentation method is dividing the whole trajectory into several small trajectories and do the model training. The proposed methods are divided into two steps: \textbf{\emph{Range Selection, and Random Selection}}. 

\begin{figure}[!ht]
  \centering
  \includegraphics[width=0.8\textwidth]{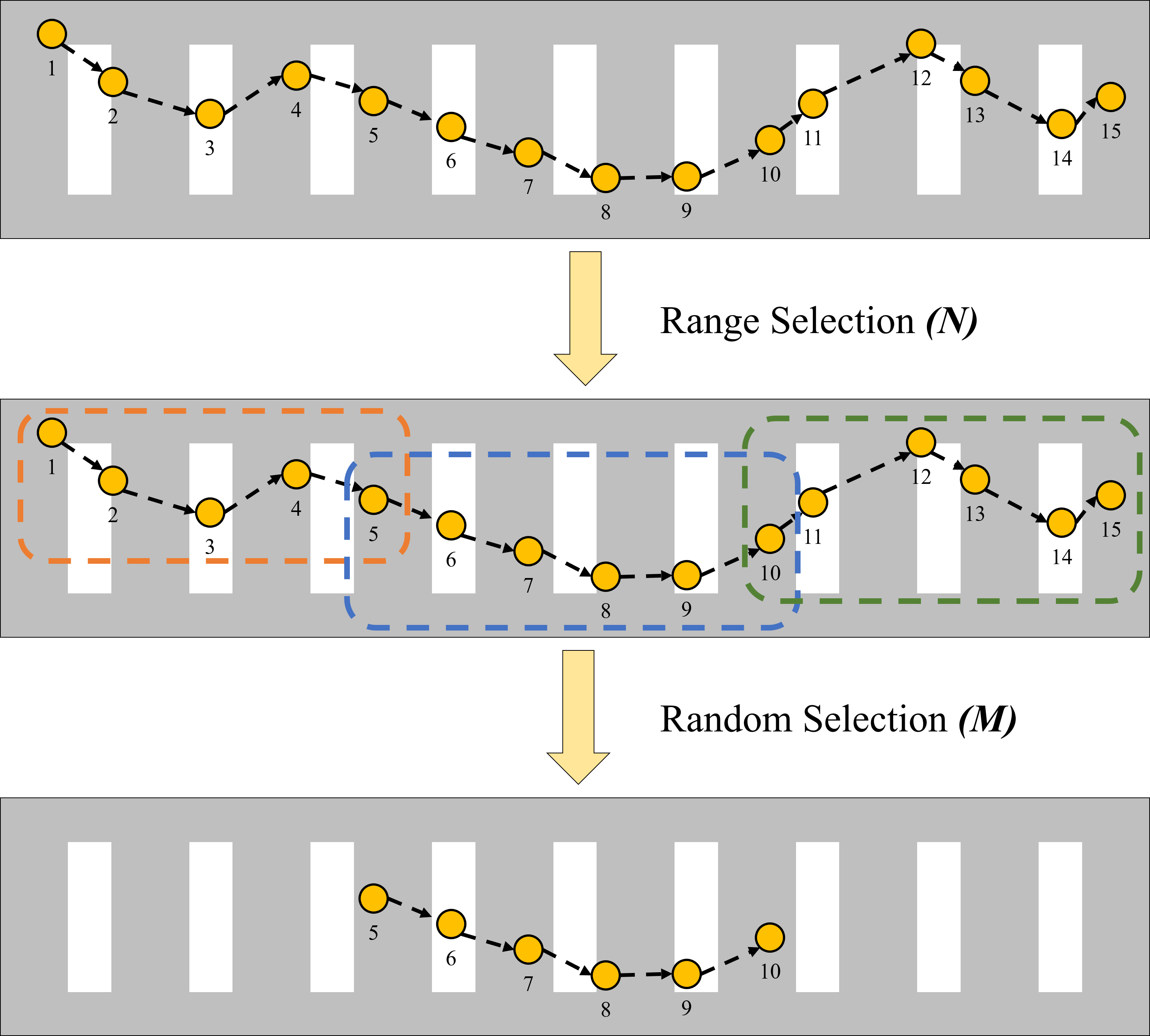}
  \caption{The framework of propose module}\label{fig:Data_augmentation}
\end{figure}

\begin{itemize}
    \item \textbf{\emph{Range Selection}}\\
In this process, we have to decide the how many points should be selected in a range. The selection range means the number of points to be decided for training. In this paper, we select the consecutive points in a certain range to ensure the model can capture the spatial and temporal feature of trajectory. For example, in the Figure~\ref{fig:Data_augmentation}, the selection range is set as 6. We show three different ranges with different colors and each range contains 6 consecutive points. For example, in the first range, which is colored as orange, we choose points from 1-5 as a sample trajectory. After range selection, we generate \textbf{\emph{N}} trajectories from a single trajectory.       

 \item \textbf{\emph{Random Selection}}\\
After range selection, we generate \textbf{\emph{N}} divided  trajectories from a single trajectory. In the next, we randomly select \textbf{\emph{M}} trajectories as input of training data. Here, the relationship between two values are $M<=N$. In the Figure~\ref{fig:Data_augmentation}, we choose $N =3$ and $M=1$.

\end{itemize}

\subsubsection{Model Introduction}
Currently, a great number of deep learning methods are promoted to solve the time series prediction since it can efficiently capture the spatio-temporal feature of big trajectory data. In order to predict the arrival time of the pedestrian accurately, several deep learning models including Deep Neural Network (DNN), Gated recurrent unit(GRU), Long short-term memory, and Transformer are performed in this paper with different ranges as the input for comparison.  

\textbf{\emph{(DNN)}}\cite{svozil1997introduction}: The Deep Neural Network (DNN) used in this paper is a sequence of feed-forward neural networks. It can efficiently figure out the relationship between predicted arrival time and the input feature without the historical sequential information of the pedestrian.

\textbf{\emph{(LSTM)}}\cite{hochreiter1997long}: The Long Short Term Memory (LSTM) is an artificial recurrent neural network designed to enable memorizing long/short-term memory. It is mainly used for time series data processing. Unlike DNN, it has feedback connections that can not only process single data points, but also entire sequences of data.

\textbf{\emph{(GRU)}}\cite{chung2014empirical}: The Gated Recurrent Unit (GRU) is a gating mechanism in recurrent neural networks. Although it has a similar structure to LSTM, it has fewer parameters, which reduces computation complexities. Empirically, GRU shows better performance than LSTM  when the amount of the data is small.

\textbf{\emph{(Transformer)}}\cite{vaswani2017attention}: The Transformer is an advanced and prominent deep learning model, which has been widely used in a variety of domains, including natural language processing, computer vision, and speech processing. The main characteristics of the model are based solely on attention mechanisms and dispense with recurrence and convolutions units entirely. In sequence to sequence (Seq2Seq) data processing, Transformer shows better performances among the proposed methods in this paper.

\subsubsection{Summary of Training Input}
Since the purpose of using deep learning models is to predict the pedestrians' arrival time, in this subsection, we further introduce the training input features in detail. The input features are mainly classified into \textbf{\emph{Time}}, \textbf{\emph{Position}}, \textbf{\emph{Size}} and \textbf{\emph{Velocity}}. 

\begin{itemize}
    \item \textbf{\emph{Time}}: The time feature consists of three parts which are the spent time, the left phase time, and the current phase. The spent time is defined as the time difference between the current and first point of the filtered trajectory. The spent time can be efficiently used in capturing the characteristics of the pedestrian, which is useful in arrival-time prediction. The left phase time is defined as the time left for the current phase to change into the next phase. The current phase represents whether the traffic light is green or not. With the current phase and the left phase time, the model can also capture the phase information to adjust its prediction results. For example, for a normal pedestrian, in a limited green time situation, the pedestrian would increase the moving velocity to cross the streets successfully. The unit used in the time feature is second.
    
    \item \textbf{\emph{Position}}: The Position of the pedestrian contains the x,y coordinate information, which aims to decide the location of the pedestrian. In this paper, the target area is divided into two different areas - the pedestrian crossing area and the vehicle area. The pedestrian area can be further divided into 4 different areas based on their location and signal phases. In Figure 5, we draw the related pedestrian crossing area into four boxes which different colors and named areas 1-4. If the point is not included in the pedestrian crossing area, we defined it to the vehicle area. In a different area, the shape and length of the road are different due to the shape of the geometry of the road. For the same pedestrian in the same scenario, the time needs to cross the road could be different in different pedestrian areas. The prediction model can extract the information from the feature for adjusting the prediction result. The unit of the coordinate is in meters.
    
    \begin{figure}[t]
    \centering
    \includegraphics[width=0.5\textwidth]{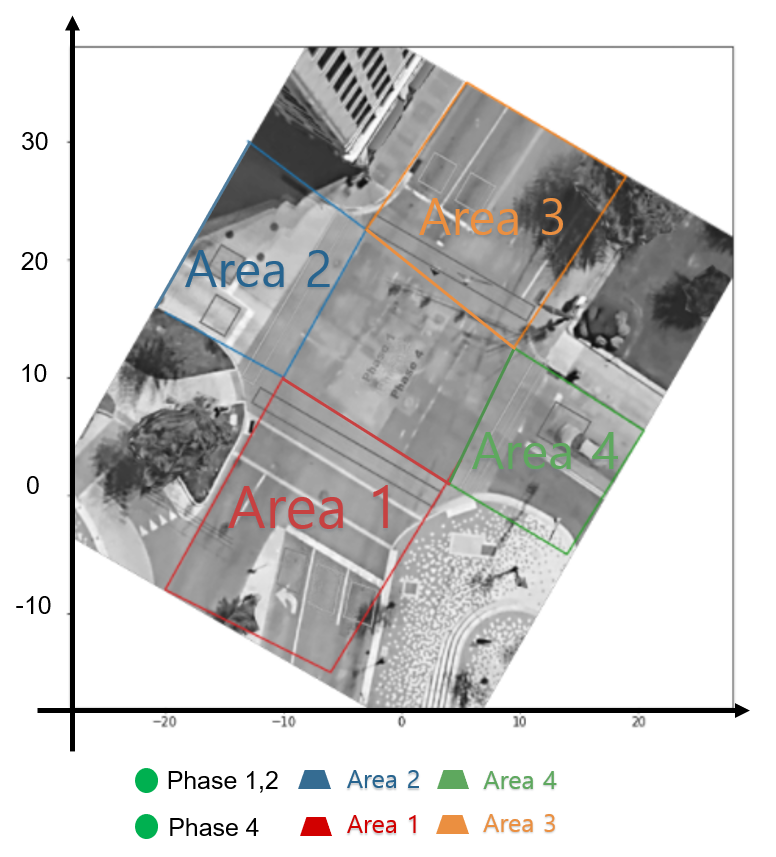}
    \caption{Area Division Map}\label{fig:3}
    \end{figure}
    
    \item \textbf{\emph{Size}}: This feature includes Bounded Box Width, Bounded Box Length, Bounded Box Height, and Yaw Angle. These four features are the basic element to decide the size of the pedestrian. The size information is one of the most important elements in defining the sub-class and also critical in developing the arrival-time prediction model.The unit of the coordinate is in meters.
    
     \item \textbf{\emph{Velocity}}: Velocity is one of the highly related features with the predicted result. Pedestrians with higher velocity can reach the destination in a shorter time. The unit of the velocity in this paper is m/s.

\end{itemize}



\section{Performance Evaluation}

\subsection{Dataset}
With the support of IEEE ITSS Technical Activities Sub-Committee ``Smart Mobility and Transportation 5.0'', The Center for Urban Informatics and Progress (CUIP) at The University of Tennessee at Chattanooga (UTC), NSF, City of Chattanooga, Ouster LiDAR, and Seoul Robotics, robust LiDAR data labeled with pedestrian, vehicle, bike and Sub-classification features of object size and velocity as well as other related features including Signal Phasing and Timing (SPaT) are provided.

The dataset provided contains two group of datasets: Competition\_Data and Demo\_Data. In each group of data contains two resource of data. One is prepossessed labeled LiDAR data. The other group of data is Signal Phasing and Timing data. 

\begin{itemize}
\item \textbf{Dataset 1} (Prepossessed Labeled LiDAR Data): Three subgroups of the dataset are provided. The data is sorted by increasing Timestamp. For each row of data, it represents one agent and its feature extracted by 3D object detection. The feature information contains Timestamp, ID, Label, Confidence, Bounded Box Position X, Bounded Box Position Y, Box Length, Box Width, Box Height, Box Yaw Angle, Velocity in the X direction, Velocity in Y direction, and Tracking\_Status. The amount of the agents in different Timestamps provided are 41130,1556694 and 54120. The time duration is on October 8, 2021, from 2:59:36 PM to 4:12:31.893 PM. There contains 3 types of agents. They are vehicles, pedestrians, and cyclists. The unit for the Timestamp, length, yaw angle and velocity are milliseconds, meter radian, and meter per second.
\item  \textbf{Dataset 2} (SPaT Data): The SPaT data contains the information of the phase scenario. The duration of the time starts from Friday, October 8, 2021, 2:00:03 PM and ends on Friday, October 8, 2021, 5:00:00 PM. The Phase Begin and Stop performance code information is provided in the SPaT data.
\item \textbf{Dataset 3} (Augmentation Dataset):  The Augmentation Dataset is generated using the Labeled Ground-Truth data by Augmentation Method. The size of the Augmentation Dataset is different according to the number of points selected in the Range Selection Method. (In the Random Selection $N=M$ for each trajectory.)
\end{itemize}

\subsection{Classification Model Result}

After Data Prepossessing and four criterion filtering, we get 96 pedestrian trajectories with 53 Normal Pedestrians, 32 Wheelchair Users, and 11 Unknown Pedestrian. Among those 11 Unknown Pedestrian, 3 of them has the similar motion information (Velocity;Criterion 1,2 ) to the Wheelchair Users while the shape information is similar to a normal pedestrian.(Height,Weight,Length,Yaw Angle;Criterion 3,4). Therefore, they supposed to be classified as the other disable people, to be more specific, an injury pedestrian. However, because of the amount of this Sub-classification is only 3. As a result, we classify the 96 trajectory into Normal Pedestrian, Wheelchair users and Unknown Pedestrian.

PCA is implemented to process the size and motion information. To reduce the effect of the error from 3D object detection and make the input feature more feasible and reliable for the classification task, only size information and motion information is chosen for the classification task. Size information includes the mean height, mean width, mean length, and the standard deviation of the pedestrian yaw angle. Motion information includes the mean velocity and the maximum velocity. With the performance of the PCA method, the visualization with label value and 2 principle components is presented. In the visualization process, the normal pedestrian is labeled as the value of 1, the wheelchair user is labeled as a value of 0 and the unknown value is labeled as a value of 2.

\begin{figure}[t]
  \centering
  \includegraphics[width=0.7\textwidth]{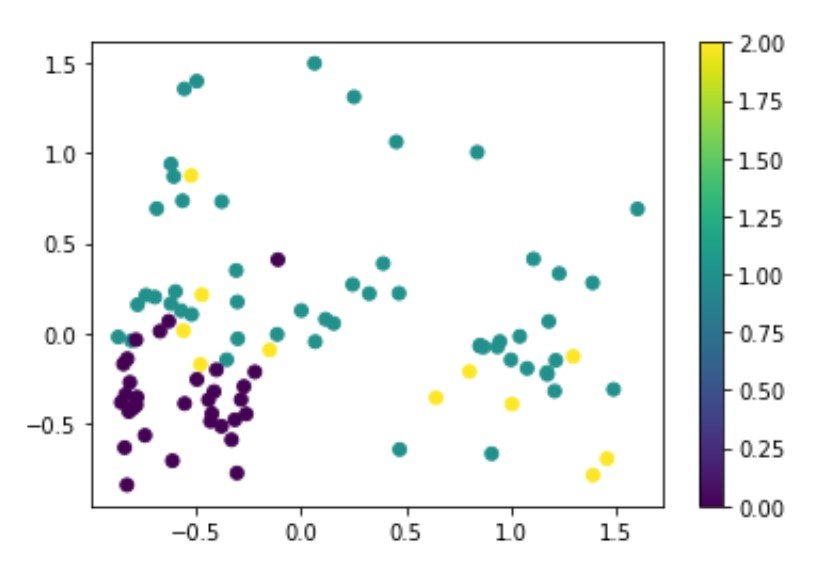}
  \caption{PCA result output}\label{fig:3}
\end{figure}
 
In the result, it shows that there is an obvious boundary between 'Normal Pedestrians' and 'Wheelchair Users' which provides that our criteria for pedestrians Sub-classification are trustworthy. Furthermore, the 'Unknown' sub-class is located at the intersection of the other two sub-classes. Therefore, the PCA output reveals the existence of the implicit relationship of the input feature with the sub-class. In order to figure out the implicit relationship between the sub-class and the size information and motion information. Then to further develop the sub-classification model, a general supervised learning method Support Vector Machine (SVM) is used.
 
To fully utilize our limited input data, we use cross-validation to ensure every observation from the original data has the chance of appearing in the training and test datasets. With the objective to train the SVM model, only trajectories labeled with 'Normal Pedestrian' and 'Wheelchair Users' are selected. The selected labeled trajectory data is split into the trained and test dataset under the setting of the split ratio as 0.8. The final results are shown that our classification model reaches around $93.3\%$ in the test sub-classifying task. The trained SVM model then is implemented on the trajectory labeled as 'Unknown' to classify them into 'Normal Pedestrian' or 'Wheelchair Users'.

\subsection{Prediction Model Result}
\subsubsection{Summary of results}
After the input data sub-classification by the model, the total number of pedestrian trajectories is 96. Table~\ref{tab:train_data} shows the detailed description of the training and test dataset. The training dataset contains 73 original trajectories and the test data includes 23 original trajectories. The hyper-parameters of the proposed four deep learning models are shown in Table~\ref{Table 2: Hyperparameters used in proposed models}.

\begin{table}[b]
    \centering
    \begin{tabular}{|l|l|l|l|l|}
\hline
\textbf{Point Number ($\emph{N}$)} & \textbf{5} & \textbf{10} & \textbf{20} & \textbf{40} \\ 
\hline
\textbf{Size of augmented trajectories (Origin: 96 trajectories)} & 9441 & 8906 & 7836  & 5716  \\   
\hline
\textbf{Augmented train dataset (Origin: 73 trajectories)}  & 7668  & 7243 & 6393 & 4707\\ \hline
\textbf{Augmented test dataset (Origin: 23 trajectories)} & 1773 & 1663  & 1443  & 1009        \\ \hline
    \end{tabular}%
    \caption{Description of training dataset}
    \label{tab:train_data}
\end{table}

\begin{table}[b]
\centering
\resizebox{0.3\textwidth}{!}{%
\begin{tabular}{|ll|}
\hline
\multicolumn{2}{|l|}{\textbf{Hyper Parameter}}          \\ \hline
\multicolumn{1}{|l|}{Batch Size}              & 30      \\ \hline
\multicolumn{1}{|l|}{Learning Rate}           & 0.00015 \\ \hline
\multicolumn{1}{|l|}{Iteration}               & 10000   \\ \hline
\multicolumn{1}{|l|}{\textbf{DNN}}            &         \\ \hline
\multicolumn{1}{|l|}{Total Layer}             & 5       \\ \hline
\multicolumn{1}{|l|}{Dimension of Layer 1}    & 256     \\ \hline
\multicolumn{1}{|l|}{Dimension of Layer 2}    & 512     \\ \hline
\multicolumn{1}{|l|}{Dimension of Layer 3}    & 256     \\ \hline
\multicolumn{1}{|l|}{Dimension of Layer 4}    & 128     \\ \hline
\multicolumn{1}{|l|}{Dimension of Layer 5}    & 64      \\ \hline
\multicolumn{2}{|l|}{\textbf{LSTM}}                     \\ \hline
\multicolumn{1}{|l|}{Hidden Layer}            & 32      \\ \hline
\multicolumn{1}{|l|}{Number of Layers}         & 2       \\ \hline
\multicolumn{2}{|l|}{\textbf{GRU}}                      \\ \hline
\multicolumn{1}{|l|}{Hidden Layer}            & 32      \\ \hline
\multicolumn{1}{|l|}{Number of Layers}         & 2       \\ \hline
\multicolumn{2}{|l|}{\textbf{Transformer}}              \\ \hline
\multicolumn{1}{|l|}{Embedding Size}          & 32      \\ \hline
\multicolumn{1}{|l|}{Number of Muti-Head Attentions}   & 4       \\ \hline
\multicolumn{1}{|l|}{Number of Layers in Encoder}    & 3       \\ \hline
\multicolumn{1}{|l|}{Number of Layers in Decoder} & 3       \\ \hline
\end{tabular}%
}
\caption{Hyperparameters used in proposed models}
\label{Table 2: Hyperparameters used in proposed models}
\end{table}

 The prediction results with different models are shown in Table~\ref{Table 3: Results of the four prediction models}. The loss function used in this experiment is the Mean Square Error (MSE). From the results, we find that the best result is performed by the GRU model with 10 points as an input which is 1.892$s^{2}$. The worse result is performed by the DNN model with 10 points as input length which is 4.879$s^{2}$. These results indicate that the error range of arrival-time prediction results is from 1.2s to 2.2s. 
 To further analyze the results, we compare the results from the perspective of different models, datasets, and training costs.

\begin{table}[!ht]
\resizebox{\textwidth}{!}{%
\begin{tabular}{|c|cc|cc|cc|cc|}
\hline
\textbf{Model}       & \multicolumn{2}{c|}{\textbf{DNN}}                     & \multicolumn{2}{c|}{\textbf{LSTM}}                    & \multicolumn{2}{c|}{\textbf{GRU}}                     & \multicolumn{2}{c|}{\textbf{Transformer}}             \\ \hline
\textbf{Result Type} & \multicolumn{1}{c|}{\textbf{MSE LOSS}} & \textbf{TIME} & \multicolumn{1}{c|}{\textbf{MSE LOSS}} & \textbf{TIME} & \multicolumn{1}{c|}{\textbf{MSE LOSS}} & \textbf{TIME} & \multicolumn{1}{c|}{\textbf{MSE LOSS}} & \textbf{TIME} \\ \hline
\textbf{5}           & \multicolumn{1}{c|}{4.137}            & 3h 56m 41s    & \multicolumn{1}{c|}{3.044}            & 5h 0m 19s     & \multicolumn{1}{c|}{3.178}            & 4h 59m 36s    & \multicolumn{1}{c|}{3.863}            & 16h 22m 20s   \\ \hline
\textbf{10}          & \multicolumn{1}{c|}{2.933}            & 5h 55m 41s    & \multicolumn{1}{c|}{\textbf{1.939}}   & 9h 54m 9s     & \multicolumn{1}{c|}{\textbf{1.892}}   & 9h 23m 25s    & \multicolumn{1}{c|}{2.889}            & 20h 6m 18s    \\ \hline
\textbf{20}          & \multicolumn{1}{c|}{4.879}            & 7h 1m 27s     & \multicolumn{1}{c|}{3.898}            & 7h 44m 10s    & \multicolumn{1}{c|}{3.674}            & 7h 43m 1s     & \multicolumn{1}{c|}{2.67}             & 18h 32m 7s    \\ \hline
\textbf{40}          & \multicolumn{1}{c|}{4.052}            & 5h 55m 55ms   & \multicolumn{1}{c|}{3.874}            & 6h 13m 7s     & \multicolumn{1}{c|}{4.352}            & 6h 12m 59s    & \multicolumn{1}{c|}{3.224}            & 18h 32m 52s   \\ \hline
\end{tabular}%
}
\caption{Results of the four prediction models}
\label{Table 3: Results of the four prediction models}
\end{table}

\begin{itemize}

\item\textbf{Training models}: 
From the perspective of models, the Transformer has stable and better performances on the four different datasets. On the contrary, The LSTM and GRU have a better performance when the number of the points of the training dataset is smaller and equal to 10. When the training data consists of ten-point trajectories, both LSTM and GRU reach their best performance. When the number of points in the training trajectory is 20 and 40, the Transformer outperforms the other models. Also, among all the models, transformer has a more stable behavior in different number of points in the training trajectory. DNN has the worse performance in all the cases due to its simple structure, which is not beneficial for capturing the time-series information. 

\item\textbf{Training dataset}: 
From the perspective of training datasets, the ten-point trajectory-based training data outperforms among other datasets in proposed deep learning models except for Transformer. These results indicate that if the time horizon of historical data is closed to 1s (one point is around 0.1second), we can predict the arrival time of the pedestrians in an accurate manner. From the results, we can also indicate that although the longer time horizon of historical data can provide more information in model training, these input data reduce the model performance. To be more specific, some input data is not utilized or suitable in model training, resulting in decreasing in model performance.

\item\textbf{Training costs}: 

From the perspective of training costs, the DNN-based prediction model shows the lowest training cost. The reason behind this is that the DNN has the simplest structure among the four proposed models. Conversely, the training cost of the Transformer is the highest (more than 3 times than DNN) due to its complex structures. However, if we consider the model prediction accuracy and training cost, we can find that the lower training costs have lower prediction results. Therefore, when we choose the model for arrival-time prediction, if we focus more on the final accuracy, we can use a more complex model. Otherwise, if we want to consider both performance and result accuracy, LSTM or GRU can be one of the good options.  
\end{itemize}

\subsubsection{Influence of sub-classification information}

\begin{figure}[!ht]
  \centering
  \includegraphics[width=1\textwidth]{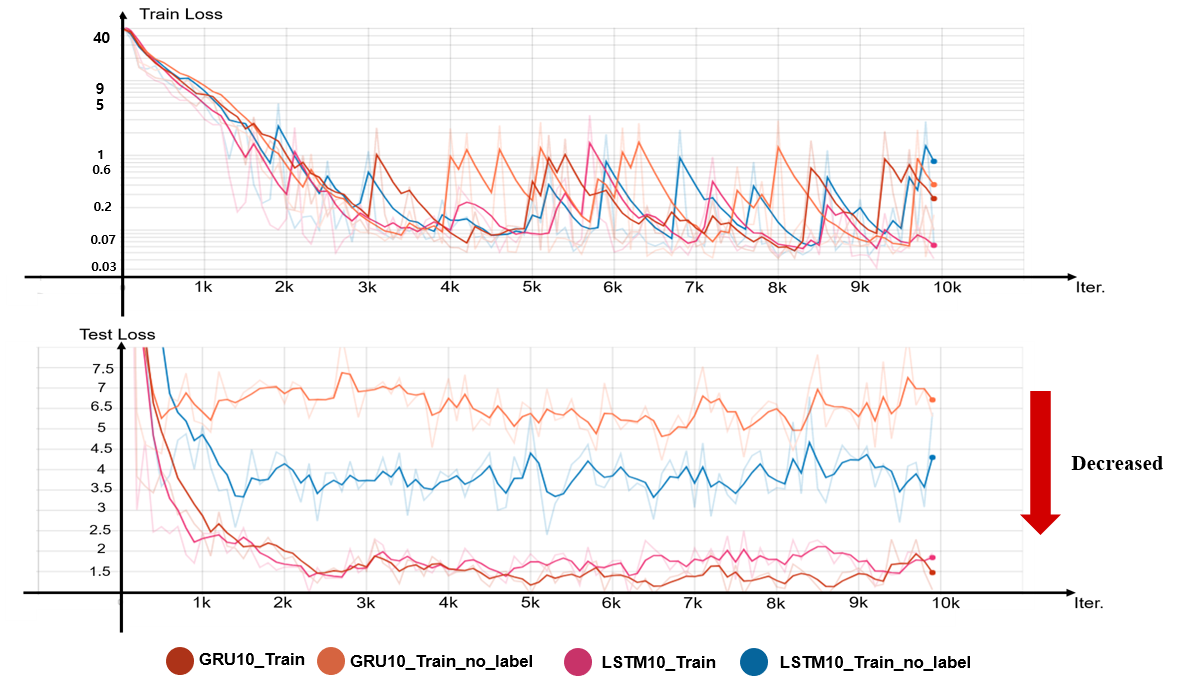}
  \caption{Test Loss with and without Label (LSTM and GRU)}\label{fig:test loss with and without label}
\end{figure}

In this part, we try to analyze the effect of sub-classification information on the final prediction results. The corresponding results are shown in Table~\ref{tab4:label_nonlabel_resutls}. In this experiment, we choose the datasets which consist of the ten-point and twenty-point trajectories since they show better and stable performances than the others. From the results, it is significant to indicate that the training dataset with sub-classification information shows better performance in the four proposed deep learning models. Specifically, the GRU trained by ten-point-based trajectories shows the biggest differences before and after adding sub-classification information in the training. The corresponding training curves are shown in Figure~\ref{fig:test loss with and without label}.

\begin{table}[!ht]
\centering
\resizebox{0.8\textwidth}{!}{%
\begin{tabular}{|c|cc|cc|cc|cc|}
\hline
MSE   & \multicolumn{2}{c|}{DNN}                   & \multicolumn{2}{c|}{LSTM}                  & \multicolumn{2}{c|}{GRU}                   & \multicolumn{2}{c|}{Transformer}           \\ \hline
Sub-classification & \multicolumn{1}{c|}{Without} & With & \multicolumn{1}{c|}{Without} & With & \multicolumn{1}{c|}{Without} & With & \multicolumn{1}{c|}{Without} & With \\ \hline
10    & \multicolumn{1}{c|}{4.955}       & 2.933   & \multicolumn{1}{c|}{4.330}       & 1.939   & \multicolumn{1}{c|}{5.495}       & 1.892   & \multicolumn{1}{c|}{4.915}       & 2.889   \\ \hline
20    & \multicolumn{1}{c|}{5.503}       & 4.879   & \multicolumn{1}{c|}{3.927}       & 3.898   & \multicolumn{1}{c|}{3.783}       & 3.674   & \multicolumn{1}{c|}{3.090}       & 2.670   \\ \hline
\end{tabular}%
}
\caption{Prediction results with the absence of sub-classification information in the training dataset}
\label{tab4:label_nonlabel_resutls}
\end{table}

Among all the models in the training, the LSTM and GRU models trained with a ten-point-based trajectory training dataset are chosen since they show better performance on the accuracy of the arrival time prediction model. In this experiment, we choose to use two different training sets: one is with sub-class information, and one without sub-class information.
The related result is shown in Figure~\ref{fig:test loss with and without label}. The results significantly reveal that the prediction model performs better in the training dataset with sub-classifying results. The newly developed model decreases differences between ground-truth and predicted arrival time around 2 seconds.  

\begin{figure}[!ht]
  \centering
  \includegraphics[width=1\textwidth]{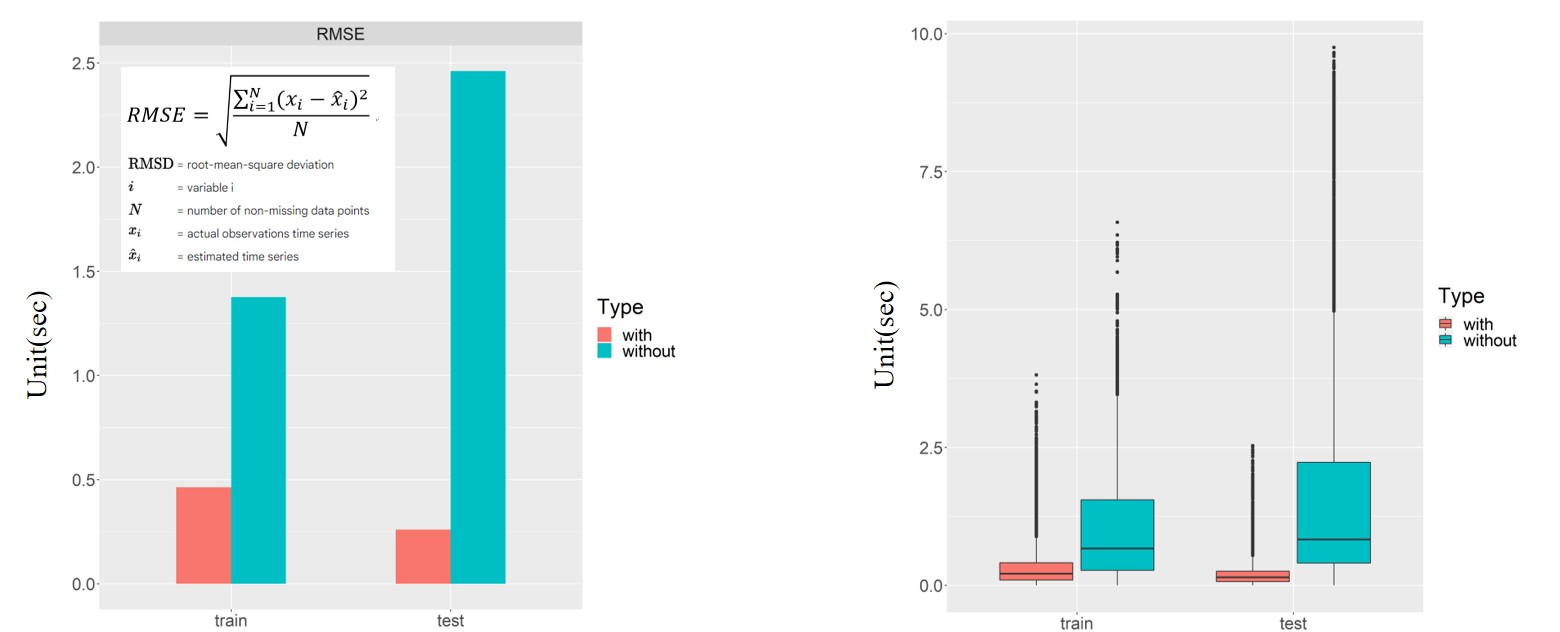}
  \caption{The box plot of final results RMSE  }\label{fig:RMSE}
\end{figure}

To further analyze the results and explore the reason for the drop in figure~\ref{fig:test loss with and without label}, we use box plots with Root Mean Square Error (RMSE). MSE is highly biased for higher values. while RMSE is better in terms of reflecting performance when dealing with large error values. RMSE is also more useful when lower residual values are preferred. The experiment is done with the best performance model which is GRU with 10 points in one trajectory. The results are shown in Figure~\ref{fig:RMSE}. From the results, we can indicate that although the training results without considering the sub-classification information have a better performance, the test results show that the range of loss in both cases is wide. It reveals that without sub-classification information, the prediction model shows less feasibility and is not robust to some specific kind of trajectory. In order to figure out how the trajectory type influences the prediction results, the performance of the prediction model in detail is presented.

\begin{figure}[!ht]
  \centering
  \includegraphics[width=1\textwidth]{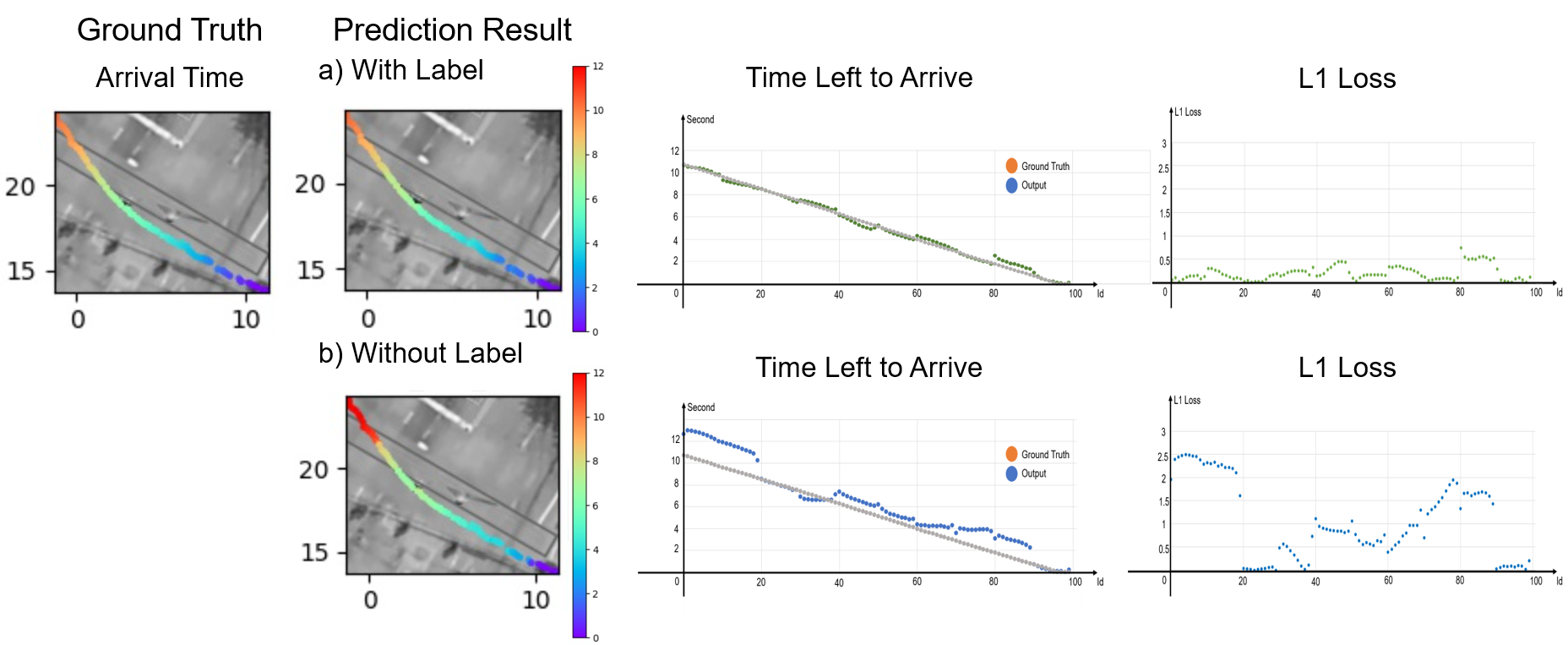}
  \caption{ID 354573 Prediction Result Comparison}\label{fig:354573}
\end{figure}

To present the performance of the prediction model in detail, we select ID 354573 as an example. Figure~\ref{fig:354573} shows the related result. The left three graphs visualize the arrival time extracted from ground truth data, predicted results with sub-classification information as input, and predicted results without sub-classification information as input. The magnitude of the arrival time is visualized by a color bar to get a more intuitive result. The conclusion that the prediction model with consideration of the sub-classification information outperforms the one without consideration of the sub-classification information can be easily reached by comparing the left three figures. The Time Left to Arrive and L1 Figure Loss show that without the sub-classification information, the prediction results are worse when far away from the other side. In addition to that, with the sub-classification information, the prediction model performs better results when the pedestrian nearly reaches his destination.

Therefore, it is necessary to consider the sub-classification feature for accurate arrival-time model development.

\subsubsection{Discussion}
The smallest difference between predicted and ground-truth arrival time reaches 1.2s is an acceptable value since the buffer time in traffic phase design is set from 3 to 5 seconds. To further improve the performance of our proposed models, we try to analyze the deficiencies of our approaches. One of the biggest problems of our model is weak in predicting the arrival time of abnormal pedestrians. To be more specific, in the trajectory with ID 371401, the pedestrian waits until the traffic phase turns green in the cross-walk road area. Since the performed data augmentation method is based on the original dataset, it cannot create pedestrian trajectories that have new features. 
In other words, due to the data sparsity problem, even one or two of the abnormal trajectories in the training or testing dataset can affect the final results. After removing abnormal trajectories, the loss among all the models is more or less decreased. This error could be minimized in 2 ways. One is using a larger dataset that contains more cases of that abnormal pedestrians. The other is performing cross-validation on the prediction model.




\section{Conclusion}
This study presented a framework for developing a novel deep learning-based model for disabling sub-classification and arrival time prediction with the prepossessed data generated by 3D object detection in traffic intersections. One of the biggest challenges in this paper is the lack of ground truth data for sub-classification. Therefore, to solve the ground-truth data problems, this study proposes four criteria to classify the overall pedestrian into the Normal pedestrian and Wheelchair users. The feasibility and reliability of our proposed criteria are verified by using PCA.  Then, SVM is used to develop a high-performing sub-classification model. The result shows around 93.3\%accuracy in the sub-classifying test. To further search for a method for the safety of disabled people, we compare several typical time series prediction models for arrival time prediction tasks. As a result, it shows that GRU with 10 points, in other words, 1 second's time horizon input, reaches the best performance with the accuracy of 1.2 seconds. Then, we analyze the effect of the existence of sub-classification information in the training dataset. The results reveal it is beneficial to include the stated information to further improve the prediction accuracy.

In the dataset, some pedestrians with violation behaviors samples are found. Violation behavior is highly relative to the nearby traffic situation (whether a vehicle is passing the road at that moment). Violation behavior shows abnormal behavior compared with normal pedestrian and wheelchair users. In this paper, the provided dataset is not adequate to consider the violation behavior. A more comprehensive model could be trained with a larger dataset to consider the influence of the violation scenario on arrival left time prediction.In addition, the calibration points provided were not accurate enough to match the background. In this paper, manually matching is done with the reference of the demon dataset. By that, some loss on the position is inevitable.

In this paper, four proposed criterion is provided to create an ideal sub-classification feature for the supervised learning method. However, the reliability of the ground-truth data has a great impact on the performance of the sub-classification model. One of the solutions is to use video data in the cross-section to sub-classify the pedestrian for providing a more reliable data source for the sub-classification model. There are several directions in which the current study can be extended to further improve the performance of both the Sub-classification Model and Time prediction Model. One extension of this work is processed with a larger data set. With a larger dataset, the over-fitted issue due to the abnormal behavior could be solved. Besides that, a larger data set will save the time cost on cross-validation and manual labeling. The SVM model could be exchanged to another deep neural network if the ground true sub-classification feature is also provided in the dataset. In that way, a more robust model can be generated. What's more, With both the classification model and the time prediction model being differentiated, an end-to-end model can be established to do both sub-class classification and arrival time prediction in one model at one time.

As one of the contributions of this paper and the requirement of the challenge, with the comparison between the predicted arrival time and the left green phase time, whether the pedestrian can cross the road can be defined. With this information, feasible implementation on the phase time can be performed to significantly improve the traffic safety at intersections by infrastructure. 






\newpage
\printcredits

\bibliographystyle{unsrtnat}
\bibliography{trb_template}

\begin{thebibliography}{18}
\providecommand{\natexlab}[1]{#1}

\bibitem[{Peden(2005)}]{peden2005global}
Peden, M., Global collaboration on road traffic injury prevention.
  \emph{International journal of injury control and safety promotion}, Vol.~12,
  No.~2, 2005, pp. 85--91.

\bibitem[{Bernhardt and Kockelman(2021)}]{bernhardt2021analysis}
Bernhardt, M. and K.~Kockelman, An analysis of pedestrian crash trends and
  contributing factors in Texas. \emph{Journal of Transport \& Health},
  Vol.~22, 2021, p. 101090.

\bibitem[{for Statistics and Analysis(2019)}]{national20192018}
for Statistics, N.~C. and Analysis, \emph{2018 fatal motor vehicle crashes:
  Overview}, 2019.

\bibitem[{WHO(????)}]{timmurphy.org}
WHO, \emph{Road traffic injuries}. [EB/OL], ????,
  \url{https://www.who.int/news-room/fact-sheets/detail/road-traffic-injuries}
  Accessed February 7, 2020.

\bibitem[{of~Transportation(????)}]{Traffic_Safety}
of~Transportation, U., \emph{TRAFFIC SAFETY FACTS 2012}. [EB/OL], ????,
  \url{https://crashstats.nhtsa.dot.gov/Api/Public/ViewPublication/812032}
  Accessed DEC 9, 2021.

\bibitem[{Coutinho et~al.(2013)Coutinho, Neto, and
  Perna}]{coutinho2013determination}
Coutinho, A., F.~Neto, and C.~Perna, Determination of normative values for 20
  min exercise of wheelchair propulsion by spinal cord injury patients.
  \emph{Spinal Cord}, Vol.~51, No.~10, 2013, pp. 755--760.

\bibitem[{Richter(2001)}]{richter2001effect}
Richter, W., The effect of seat position on manual wheelchair propulsion
  biomechanics: a quasi-static model-based approach. \emph{Medical engineering
  \& physics}, Vol.~23, No.~10, 2001, pp. 707--712.

\bibitem[{Wold et~al.(1987)Wold, Esbensen, and Geladi}]{wold1987principal}
Wold, S., K.~Esbensen, and P.~Geladi, Principal component analysis.
  \emph{Chemometrics and intelligent laboratory systems}, Vol.~2, No. 1-3,
  1987, pp. 37--52.

\bibitem[{Noble(2006)}]{noble2006support}
Noble, W.~S., What is a support vector machine? \emph{Nature biotechnology},
  Vol.~24, No.~12, 2006, pp. 1565--1567.

\bibitem[{Jin et~al.(2021)Jin, Kim, Yeo, and Choi}]{jin2021transformer}
Jin, Z., J.~Kim, H.~Yeo, and S.~Choi, Transformer-based Map Matching Model with
  Limited Ground-Truth Data using Transfer-Learning Approach. \emph{arXiv
  preprint arXiv:2108.00439}, 2021.

\bibitem[{Travis and Bevly(2008)}]{travis2008trajectory}
Travis, W. and D.~M. Bevly, Trajectory duplication using relative position
  information for automated ground vehicle convoys. In \emph{2008 IEEE/ION
  Position, Location and Navigation Symposium}, IEEE, 2008, pp. 1022--1032.

\bibitem[{Chen et~al.(2011)Chen, Shen, and Zhou}]{chen2011discovering}
Chen, Z., H.~T. Shen, and X.~Zhou, Discovering popular routes from
  trajectories. In \emph{2011 IEEE 27th International Conference on Data
  Engineering}, IEEE, 2011, pp. 900--911.

\bibitem[{Wang et~al.(2021)Wang, Liu, Lu, and Yang}]{wang2021large}
Wang, X., X.~Liu, Z.~Lu, and H.~Yang, Large Scale GPS Trajectory Generation
  Using Map Based on Two Stage GAN. \emph{Journal of Data Science}, Vol.~19,
  No.~1, 2021, pp. 126--141.

\bibitem[{Choi et~al.(2021)Choi, Kim, and Yeo}]{choi2021trajgail}
Choi, S., J.~Kim, and H.~Yeo, TrajGAIL: Generating urban vehicle trajectories
  using generative adversarial imitation learning. \emph{Transportation
  Research Part C: Emerging Technologies}, Vol. 128, 2021, p. 103091.

\bibitem[{Svozil et~al.(1997)Svozil, Kvasnicka, and
  Pospichal}]{svozil1997introduction}
Svozil, D., V.~Kvasnicka, and J.~Pospichal, Introduction to multi-layer
  feed-forward neural networks. \emph{Chemometrics and intelligent laboratory
  systems}, Vol.~39, No.~1, 1997, pp. 43--62.

\bibitem[{Hochreiter and Schmidhuber(1997)}]{hochreiter1997long}
Hochreiter, S. and J.~Schmidhuber, Long short-term memory. \emph{Neural
  computation}, Vol.~9, No.~8, 1997, pp. 1735--1780.

\bibitem[{Chung et~al.(2014)Chung, Gulcehre, Cho, and
  Bengio}]{chung2014empirical}
Chung, J., C.~Gulcehre, K.~Cho, and Y.~Bengio, Empirical evaluation of gated
  recurrent neural networks on sequence modeling. \emph{arXiv preprint
  arXiv:1412.3555}, 2014.

\bibitem[{Vaswani et~al.(2017)Vaswani, Shazeer, Parmar, Uszkoreit, Jones,
  Gomez, Kaiser, and Polosukhin}]{vaswani2017attention}
Vaswani, A., N.~Shazeer, N.~Parmar, J.~Uszkoreit, L.~Jones, A.~N. Gomez,
  {\L}.~Kaiser, and I.~Polosukhin, Attention is all you need. In \emph{Advances
  in neural information processing systems}, 2017, pp. 5998--6008.

\end{thebibliography}

\end{document}